\documentclass[lettersize,journal]{IEEEtran}
\usepackage{amsmath,amsfonts}
\pdfoutput=1
\usepackage{algorithmic}
\usepackage{algorithm}
\usepackage{array}
\usepackage[caption=false,font=normalsize,labelfont=sf,textfont=sf]{subfig}
\usepackage{textcomp}
\usepackage{stfloats}
\usepackage{url}
\usepackage{verbatim}
\usepackage{graphicx}
\usepackage{cite}
\usepackage{multirow}
\usepackage{bbding}
\begin{document}

\title{Wavelet-Like Transform-Based Technology in Response to the Call for Proposals on Neural Network-Based Image Coding}
\author{
	Cunhui Dong, 
	Haichuan Ma, 
	Haotian Zhang, 
	Changsheng Gao, 
	Li Li, \IEEEmembership{Member, IEEE,}
	and Dong Liu, \IEEEmembership{Senior Member, IEEE}
	\thanks{
				C. Dong, H. Ma, H. Zhang, C. Gao, L. Li, and D. Liu are with the CAS Key Laboratory of Technology in Geo-Spatial Information Processing and Application System, University of Science and Technology of China, Hefei 230027, China~(e-mail: dongcunh@mail.ustc.edu.cn; 
				hcma@mail.ustc.edu.cn; 
				zhanghaotian@mail.ustc.edu.cn; 
				changshenggao@mail.ustc.edu.cn; 
				lil1@ustc.edu.cn; 
				dongeliu@ustc.edu.cn
				). 
				\emph{(Corresponding author: Li Li)}
				
	}
}
%



\maketitle

\begin{abstract}
Neural network-based image coding has been developing rapidly since its birth. Until 2022, its performance has surpassed that of the best-performing traditional image coding framework -- H.266/VVC. Witnessing such success, the IEEE 1857.11 working subgroup initializes a neural network-based image coding standard project and issues a corresponding call for proposals (CfP). In response to the CfP, this paper introduces a novel wavelet-like transform-based end-to-end image coding framework -- iWaveV3. iWaveV3 incorporates many new features such as affine wavelet-like transform, perceptual-friendly quality metric, and more advanced training and online optimization strategies into our previous wavelet-like transform-based framework iWave++. While preserving the features of supporting lossy and lossless compression simultaneously, iWaveV3 also achieves state-of-the-art compression efficiency for objective quality and is very competitive for perceptual quality. As a result, iWaveV3 is adopted as a candidate scheme for developing the IEEE Standard for neural-network-based image coding.
\end{abstract}


\section{Introduction}

\IEEEPARstart{I}{n} the past decade, deep neural networks (DNN) have achieved tremendous development and shown great potential in many fields. Especially, in the field of image coding, DNN-based image coding methods have achieved great success and surpassed traditional image coding methods.
Roughly speaking, DNN-based image coding methods can be divided into two categories. One category is to design DNN-based modules, which are then integrated into the traditional image coding framework. This category is often referred to as deep tools. The other category is to design an image coding framework using DNNs solely, which is often referred to as end-to-end frameworks.
In this paper, we mainly focus on end-to-end frameworks.

In 2016, Balle $et$ $al$. introduced the first end-to-end framework \cite{balle2016end} using DNNs. The framework is mainly composed of three modules including transform, quantization, and entropy coding, which is followed by most recent researches. 
The transform usually uses an auto-encoder, which transforms the image from the high-redundant pixel domain to a compact latent domain. 
The quantization is used to round the float latent representation. As the quantization is non-differentiable, the uniform noise is usually used to simulate quantization during training to achieve gradient back-propagation.
The entropy coding initially uses the factorized model to encode the representation. Later the hyperprior and autoregressive context models are introduced to further improve compression performance.
In just two years, the compression performance of end-to-end frameworks surpassed that of BPG.

The pioneering work from Balle $et$ $al$. shows great potential of end-to-end frameworks. Many researchers pour into this field to further optimize the performance of end-to-end frameworks from a variety of aspects.  
In terms of transform, some work introduces lossless transform such as wavelet-like transform \cite{ma2022iwave++} and normalizing flow\cite{zhang2021ivpf, helminger2020lossy} to achieve lossy and lossless coding simultaneously. 
In terms of quantization, the soft-to-hard quantization \cite{agustsson2020universally} and soft-then-hard quantization \cite{guo2021soft} are proposed to reduce the quantization mismatch between training and test.
Moreover, some frameworks \cite{agustsson2019generative, mentzer2020high} have been designed to significantly improve the perceptual qualities of reconstructed images. 
For the problem that DNN parameters are usually trained on a large dataset but not optimized for a specific image, some online optimization work \cite{djelouah2019content, yang2020improving} has been proposed. 
After these major improvements, the compression performance of end-to-end frameworks has significantly surpassed that of the best traditional image coding framework VVC in terms of both objective and perceptual quality.

Witnessing such success of end-to-end image coding frameworks, various standardization groups have started developing end-to-end image coding standards to promote the application of end-to-end image coding. 
In 2021, JPEG AI started a learning-based image coding standard targeting both human visualization and computer vision tasks. 
In the same year, IEEE 1857.11 working subgroup issued a call for proposals (CfP) for neural network-based image coding standard for human visualization.


In response to the IEEE 1857.11 CfP, this paper introduces a novel wavelet-like transform-based end-to-end image coding framework -- iWaveV3.
iWaveV3 mainly includes transform, quantization, entropy coding, and post-processing modules. 
For the transform module, additive wavelet-like transform, affine wavelet-like transform, or CDF 5/3 wavelet transform can be selected according to the coding configuration. 
For the quantization module, the transform coefficients are first divided by a quantization step and then quantized to integers by rounding.
The entropy coding module efficiently codes quantized coefficients through an autoregressive context model.
The post-processing module is used to reduce the distortion caused by quantization. 
In addition, to further improve its compression performance, we introduce advanced training and online optimization strategies. 
Furthermore, a perceptual-friendly quality metric is designed to improve the perceptual qualities of reconstructed images.  
Experimental results show that for lossy compression, iWaveV3 achieves state-of-the-art compression efficiency for objective quality and is also very competitive for perceptual quality. For lossless compression, it is comparable to FLIF\cite{sneyers2016flif}. 
Therefore, it is adopted as a candidate scheme for developing the IEEE Standard for Neural Network-Based Image Coding (No. 1857.11).

The iWaveV3 presented in this paper is built upon iWave++ \cite{ma2020end}.
While preserving the feature of simultaneously supporting loss and lossy compression, it incorporates many new features into iWave++ to improve its training stability and compression performance, as shown in Table \ref{diffence}. First, iWaveV3 uses a more flexible affine wavelet-like transform. Second, more advanced training and quantization strategies are introduced. Third, in addition to the objective metric, iWaveV3 also adds a perceptual metric to improve the perceptual qualities of reconstructed images. Finally, online optimization technology is incorporated to improve compression performance further.

\begin{table*}[]
	\caption{Differences between the iWave++\cite{ma2020end} and the iWaveV3 in this paper. }
	\begin{center}
	\begin{tabular}{c|c|c}
		\hline
		& iWave++\cite{ma2020end}  & iWaveV3          \\ \hline
		Wavelet-like transform & Additive transform & Additive and affine transform \\
		Quantization     & STE\cite{theis2017lossy} quantization & Soft-to-hard and soft-then-hard quantization        \\
			Metric   & Objective metric & Objective and perceptual metric         \\
		Online optimization  & No & Yes         \\
	 \hline
	\end{tabular}
\end{center}

\label{diffence}
\end{table*}

The remainder of this paper is organized as follows. Section \ref{sec_related_work} gives a brief review of related work, and Section \ref{method} introduces the overall framework and details of each module. Section \ref{experment} presents the experimental configuration and the performance of iWaveV3. Finally, Section \ref{conclusion} draws the conclusion.

\section{Related Work}
\label{sec_related_work}
Since the pioneering work \cite{balle2016end} was proposed for end-to-end image compression, learning-based image compression has been intensively investigated recently. In this paper, progresses made in transform, quantization, perceptual quality, and online optimization are introduced. 
\subsection{Transform}
Transform is the most powerful way of reducing spatial redundancy and is adopted in almost all conventional image compression solutions. However, the transforms in conventional image coding standards, such as discrete cosine transform (DCT) in JPEG [14], wavelet transform in JPEG2000 [15], etc., are all linear, resulting in a lack of ability to capture the nonlinear correlation between pixels. To address this problem, various nonlinear transforms are proposed in learned image compression. Unlike the transform in traditional image coding standards where the transform is invertible and thus lossless, transforms in learned image compression methods can be grouped into two categories: lossy transform and lossless transform.
\subsubsection{Lossy Transform}
The lossy transform is fulfilled by transforming the input into a small latent with a lower dimension. The redundancy among the input is reduced while part of the information contained in the input is dropped. Balle $et$ $al$. \cite{balle2016end} first proposed the CNN-based transform, termed analysis transform and synthesis transform, in an end-to-end image compression framework. With the generalized division normalization (GDN) \cite{balle2015density}, the proposed analysis and synthesis transforms were equipped with nonlinearity. 
Many subsequent learned image compression methods \cite{theis2017lossy,balle2018variational,mentzer2018conditional,cheng2020learned,nakanishi2019neural,rippel2017real,liu2019non} follow the auto-encoder structure. Methods in \cite{theis2017lossy,mentzer2018conditional} integrated residual blocks into auto-encoder to enlarge the nonlinearity of transform. A multiscale structure was proposed to extend the network capacity in \cite{nakanishi2019neural,rippel2017real}. Attention mechanism based on neural transformer unit was utilized in \cite{lu2021transformer,bai2022towards,zhu2022transformer,jeny2022efficient} to improve the compression efficiency.
\subsubsection{Lossless Transform}
Due to the characteristic of outputting fewer dimensions, auto-encoder-based transform is destined lossy. To achieve lossless image compression, Ma $et$ $al$. proposed a trained wavelet-like transform that converts images into coefficients without any information loss in the end-to-end optimized image compression framework \cite{ma2020iwave,ma2022iwave++}. Wavelet-like transform was extended in volumetric image compression \cite{xue2022aiwave,xue2021iwave3d} given its high performance. 
In addition, researchers explore the opportunity for leveraging normalizing flows for end-to-end image compression. A normalizing flow admits a learnable bijective mapping between the original image and the implicit representation, enabling lossless image coding \cite{hoogeboom2019integer,berg2020idf++,zhang2021ivpf,ho2019compression,zhang2021iflow}. By introducing quantization, lossy image compression using normalizing flow was achieved in \cite{helminger2020lossy,ho2021anfic}.
\subsection{Quantization}
Quantization makes end-to-end optimization an intractable problem because of its non-differentiable characteristic. To resolve this problem, four kinds of approaches are proposed to approximate quantization in learned image compression. 

Since the gradient of quantization is almost zero everywhere, the most straightforward idea is to manually set a gradient for it. The STE approach \cite{bengio2013estimating} applies the identity gradient to pass through the hard rounding layer to enable backpropagation. This approach was adopted in \cite{theis2017lossy, mentzer2018conditional}. Balle $et$ $al$. proposed to apply additive uniform noise (AUN) during the training phase as a soft approximation to hard rounding in \cite{balle2016end}. At the inference phase, they directly round the latent and transmit them with entropy coding. Obviously, there is a mismatch between the training and test phases, which can be theoretically attributed to the variational relaxation of the actual rate, leading to the suboptimal rate-distortion performance. 
Annealing-based algorithms were proposed to approximate quantization \cite{agustsson2017soft,yang2020improving,agustsson2020universally}. By decreasing the value of temperature, the differentiable approximation function goes towards the shape of hard rounding gradually. 
However, the gradient goes zero or infinite when the soft approximation is close to hard rounding, making it hard to train the whole model stably. 
In \cite{guo2021soft}, they claimed that the annealing-based approaches restrict latent expressiveness. To cope with both train-test mismatch and expressiveness challenges, they proposed to combine AUN-based and hard rounding approaches as the soft-then-hard approach. In the soft-then-hard approach, the quantization is first approximated as additive uniform noise and the whole network is trained. After a while, the hard rounding is applied and only the network after quantization is optimized.
\subsection{Perceptual Quality}
Images usually serve the human eye and thus perceptual experience is an important factor in image quality evaluation. Conventional image coding standards optimize the reconstruction quality by mean squared error (MSE) which has been demonstrated not consistent with the human visual experience. Thanks to the end-to-end optimization mechanism, perceptual quality can be optimized by replacing MSE with an appropriate metric. 

It is believed that feature loss can improve the perceptual quality and they are applied in \cite{mentzer2020high, gao2022towards}. The image is first mapped into a high-dimensional space and the feature loss, such as MSE, is computed there. 
Generative adversarial networks (GAN) \cite{goodfellow2020generative} have shown their ability of generating realistic and sharper images and thus are utilized in the image compression field \cite{rippel2017real,johnson2016perceptual,santurkar2018generative,agustsson2019generative,mentzer2020high,blau2019rethinking,tschannen2018deep} for better perceptual quality. In \cite{agustsson2019generative}, the proposed method synthesizes details it cannot afford to store, obtaining visually pleasing results at bitrates where previous methods fail and show strong artifacts. By combining MSE loss, feature loss and GAN loss, Mentzer $et$ $al$. \cite{mentzer2020high} achieves stunning perceptual quality.

\subsection{Online Optimization}
Once the training finishes, the whole compression network is fixed and cannot adapt to a specific image as it does in traditional image coding standards. Considering that adaptability can improve coding efficiency dramatically, online optimization is investigated in the end-to-end image compression approaches. 

There are two kinds of online optimization approaches in the image compression field. The first one is to online train the compression network to make it more suitable for a specific image, such as the methods proposed in \cite{jiang2022online,yang2020improving}. In \cite{yang2020improving}, they identify three approximation gaps (the amortization gap, the discretization gap, and the marginalization gap) that limit the compression efficiency and propose remedies for each of these three limitations. The other approach is to modify the input image \cite{wang2021substitutional,djelouah2019content}, making it more compressible for a specific compression network. In \cite{djelouah2019content}, they introduce an iterative procedure that adapts the latent representation to the specific content while keeping the parameters of the network and the predictive model fixed. The second approach can be viewed as a preprocessing technique, while the input image is optimized directly by the gradient computed from the quality metric rather than in a heuristic way.

\section{The Proposed Method}
\label{method}
In this section, we present the iWaveV3 end-to-end image coding framework. First, we introduce the overall structure of this framework for both lossy and lossless coding. Then the transform, entropy coding, and post-processing modules are discussed. Finally, the training strategy and online optimization are introduced, respectively.
\subsection{Overview}
\begin{figure*}
	\centering
	\includegraphics[width=2\columnwidth]{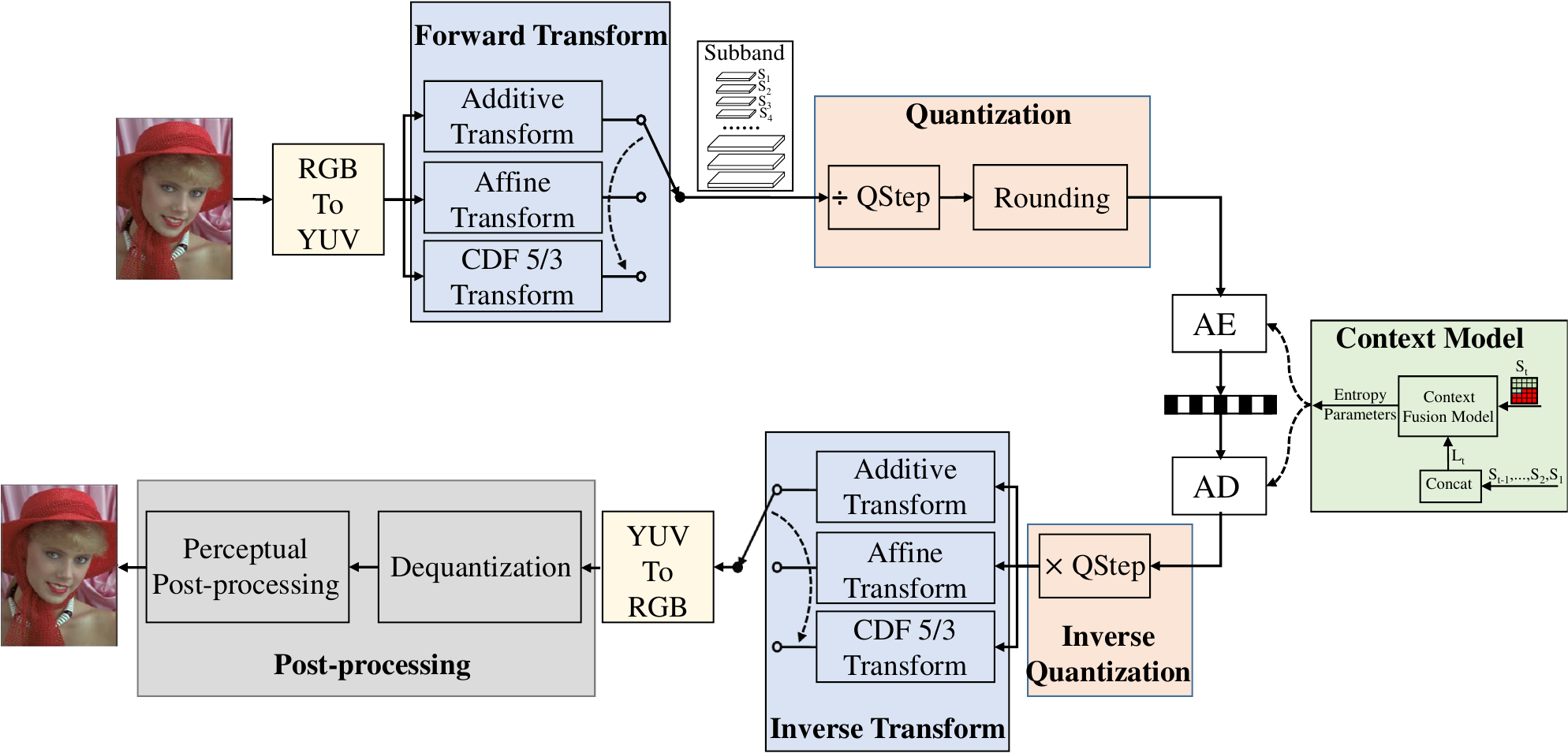} \\
	\caption{
		Overview of the proposed iWaveV3. It mainly consists of four modules. 
		The transform module offers three options, namely, the additive transform, affine transform, and CDF 5/3 wavelet transform, with the affine and additive transform implemented using CNN. 
		The quantization module initially divides the subbands by the QStep and then quantizes them to integers using rounding. 
		The entropy coding module is used to code quantized coefficients into bitstream through an autoregressive context model.
		The post-processing module is used to alleviate the quantization distortion and improve the qualities of reconstructed images. 
}
	\label{overview}
\end{figure*}
Fig.~\ref{overview} shows the overview of the proposed iWaveV3 framework. It includes color space conversion, wavelet-like transform, quantization, entropy coding, and so on. 
On the encoder side, an RGB image is first converted to a YUV image to reduce the correlation among the color channels. This enables the independent processing of Y, U, and V components.
Second, the YUV image is sent to the wavelet-like transform module to obtain several subbands. 
The transform module offers a choice of additive wavelet-like transform, affine wavelet-like transform, or CDF 5/3 wavelet transform.
Third, the transformed subband coefficients are fed into the quantization module which consists of two steps: division and rounding. QStep used in the division is learned from a DNN or set to 1 for lossy and lossless coding, respectively.
Fourth, the quantized coefficients go through the entropy coding module to generate the bitstream.
As the key of the entropy coding module, the context model obtains entropy parameters from the currently coded subband and the previously coded subbands, which are then used to derive the probability distribution of each quantized coefficient and generate the bitstream.

On the decoder side, the first step involves decoding the coefficients from the bit stream through the arithmetic decoding engine. 
Then the coefficients undergo inverse quantization by multiplying QStep which is consistent with that of the encoder. 
Subsequently, a wavelet-like inverse transform is performed on the inverse quantized coefficients to reconstruct the YUV image, and the inverse transform shares parameters with that of the forward transform. 
After that, the YUV image is converted back to an RGB image. The lossless coding process ends here while the post-processing is required for lossy coding. 

\subsection{Wavelet-like Transform}
\begin{figure}
	\centering
	\begin{tabular}{cc}
		\includegraphics[width=0.9\columnwidth]{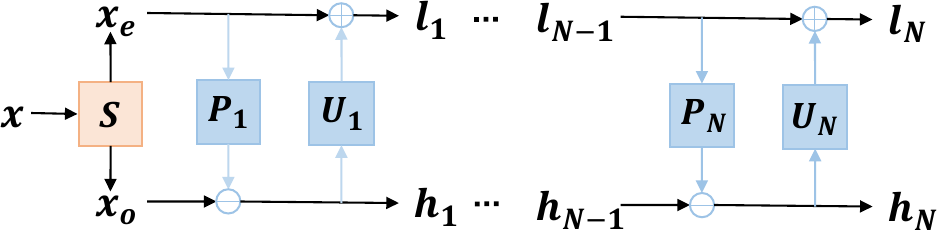} \\
		{\small (a) Additive wavelet-like transform}\\
		\includegraphics[width=0.9\columnwidth]{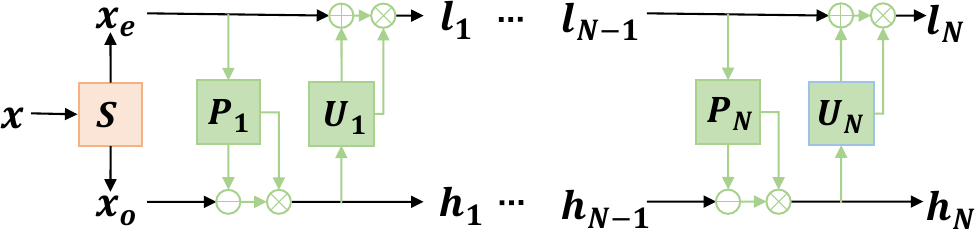} \\
		{\small (b) Affine wavelet-like transform}\\
	\end{tabular}
	
	\caption{
		The structure of additive wavelet-like transform and affine wavelet-like transform. $S$ stands for split, $P_i$ and $U_i$ stand
		for the i-th prediction unit and the i-th update unit, respectively. They are both constructed by CNN. $N$ is the number of lifting steps.}
	\label{wavelet-transform}
\end{figure}

The three types of transform in the wavelet transform module are: additive wavelet-like transform, affine wavelet-like transform, and CDF 5/3 wavelet transform \cite{christopoulos2000jpeg2000}. For lossy coding, additive wavelet-like transform or affine wavelet-like transform is used as the transform module. Affine wavelet-like transform has higher flexibility, but it is more difficult to train especially at the high bitrate. Therefore, affine wavelet-like transform is used at the low and medium bitrate, and additive wavelet-like transform is used at the high bitrate. 
For lossless coding, we use the traditional CDF 5/3 wavelet. 
The reason for using CDF 5/3 wavelet is that compared with the learnable wavelet, CDF 5/3 wavelet has much fewer parameters and is easier to train despite a little performance drop \cite{ma2020end}.



\subsubsection{\textbf{Additive Wavelet-Like Transform}}
The additive wavelet-like transform uses the lifting scheme, as illustrated in Fig.~\ref{wavelet-transform}(a). 
The transform comprises three fundamental stages: split, prediction, and update. 
For illustration, we utilize a one-dimensional signal to show the transform process.
First, the signal $x$ is split into two parts $x_e$ and $x_o$ according to the parity. Second, the prediction operation is performed to obtain $h_1$ as follows:
\begin{equation}
	\begin{aligned}
		h_1 = x_o - P_1(x_e)
	\end{aligned}
	\label{mvfor-haar-wavelet}
\end{equation}
Third, the update operation is performed to obtain $l_1$ as follows:
\begin{equation}
	\begin{aligned}
		l_1 = x_e + U_1(h_1)
	\end{aligned}
	\label{mvfor-haar-wavelet}
\end{equation}
The prediction and update operations are repeated to obtain the final $l_N$ and $h_N$.
The network structures of $P_i$ and $U_i$  use the same plain CNN architecture, as depicted in Fig.~\ref{P-U}(a).

\subsubsection{\textbf{Affine Wavelet-Like Transform}}
\begin{figure}
	\centering
	\begin{tabular}{cc}
		\includegraphics[width=0.7\columnwidth]{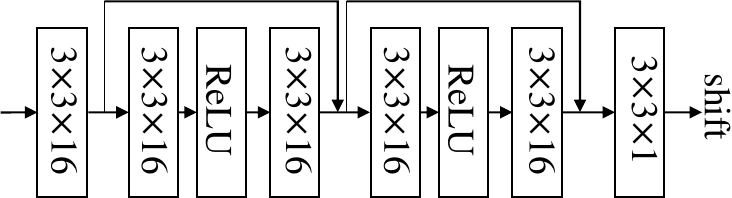} \\
		{\small (a)  $P_i$ or  $U_i$ of additive wavelet-like transform}\\
		\includegraphics[width=0.8\columnwidth]{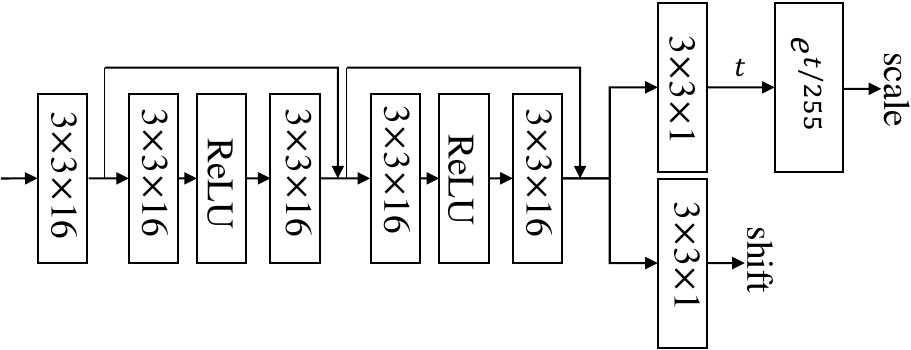} \\
		{\small (b)  $P_i$ or  $U_i$ of affine wavelet-like transform}\\
	\end{tabular}
	
	\caption{The $P_i$ or $U_i$ structure of additive wavelet-like transform and affine wavelet-like transform. The numbers like 3$\times$3$\times$16 indicate the kernel size (3$\times$3) and
		the number of channels (16). ReLU indicates the adopted
		nonlinear activation function.}
	\label{P-U}
\end{figure}

Our proposed affine wavelet-like transform is shown in Fig.~\ref{wavelet-transform}(b). Unlike the additive wavelet-like transform, which only uses addition and subtraction operations for prediction and update, the affine wavelet-like transform incorporates a scaling operation, providing greater non-linear capability.
Similar to the additive wavelet-like transform, the affine wavelet-like transform also involves three steps: split, prediction, and update. The split step is the same as that in the additive wavelet-like transform.
The prediction operation of affine wavelet-like transform is performed as follows:
\begin{equation}
	\begin{aligned}
		&shift, scale = P_1(x_e) \\
		&h_1 = scale \odot (x_o - shift)
	\end{aligned}
	\label{mvfor-haar-wavelet}
\end{equation}
and the update operation is performed as follows:
\begin{equation}
	\begin{aligned}
		&shift, scale = U_1(h_1) \\
		&l_1 = scale \odot (x_e + shift)
	\end{aligned}
	\label{mvfor-haar-wavelet} 
\end{equation}
where $\odot$ represents the matrix dot product. 
$P_i$ and $U_i$ use the same CNN and Fig. \ref{P-U}(b) shows the detailed network structure of $P_i$ or $U_i$. To stabilize the training process, the scales of $P_i$ and $U_i$ are exponentiated with $e$ to constrain the scale to positive values, as shown in Fig.~\ref{P-U}(b).

\subsubsection{\textbf{Transform for Two-Dimensional Images}}
\begin{figure}
	\centering
	\includegraphics[width=1\columnwidth]{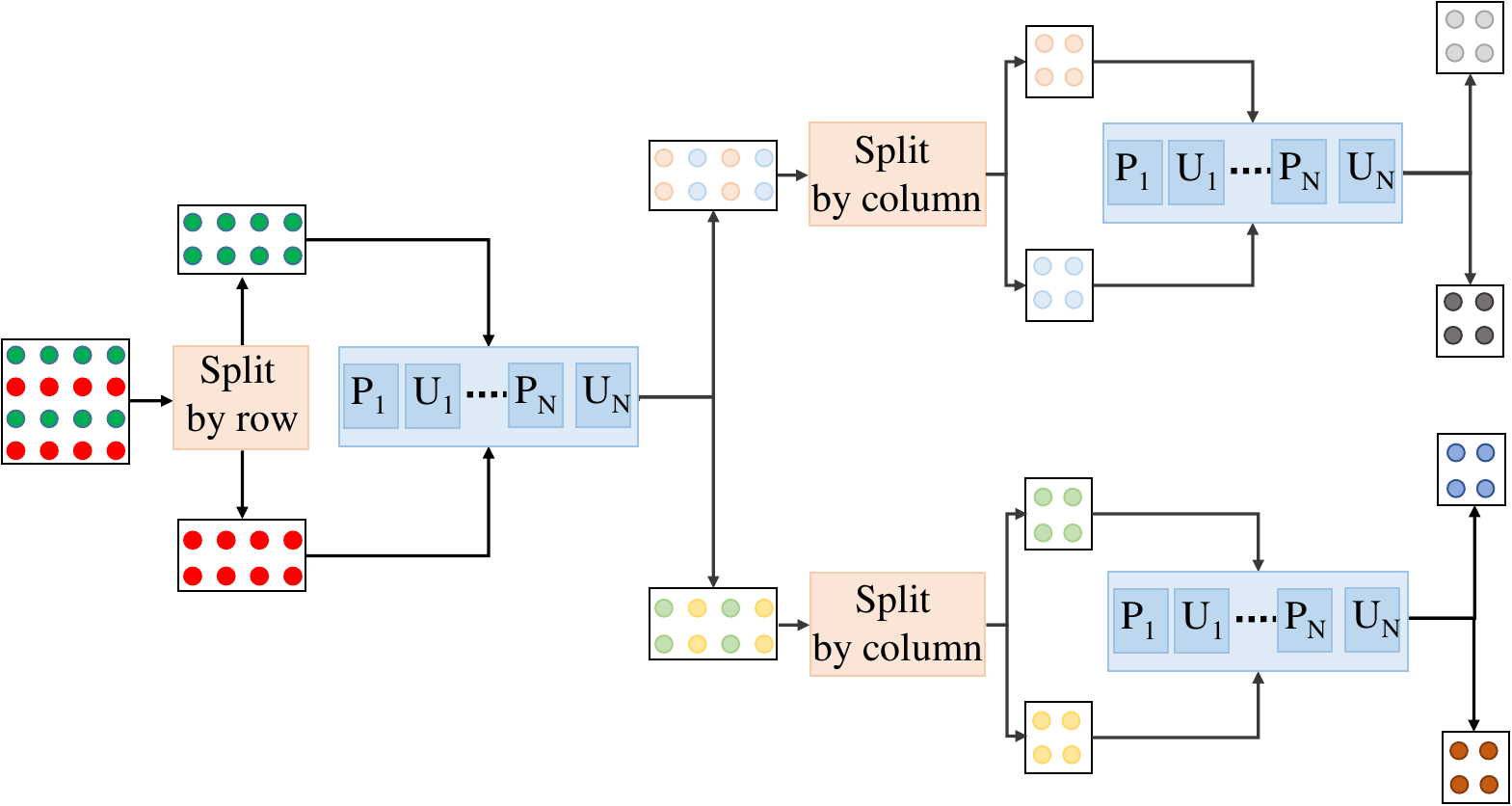} \\
	\caption{The pipeline of forward transform for two-dimensional images. The $P_i$ and $U_i$ can use that of additive wavelet-like transform, affine wavelet-like transform, or CDF 5/3 wavelet transform. }
	\label{transform}
\end{figure}
For a two-dimensional image, row-wise transform and column-wise transform are performed sequentially, as shown in Fig. \ref{transform}. The $P$ and $U$ can be selected from those of additive wavelet-like transform, affine wavelet-like transform, or CDF 5/3 wavelet transform. With transform applied, the image is converted into several subbands \{$LL_1, LH_1, HL_1, HH_1$\}. $L$ and $H$ stand for low-frequency and high-frequency, respectively. Multiscale pyramid transform is performed to further decorrelate the low-frequency signals, i.e., the next level of transform is performed on the low-frequency subband of the previous level of transform. For example, the second level of transform is performed on $LL_1$ to obtain \{$LL_2, LH_2, HL_2, HH_2$\}.


\subsection{Entropy Coding}

\begin{figure}
	\centering
	\begin{tabular}{cc}
		\includegraphics[width=0.4\columnwidth]{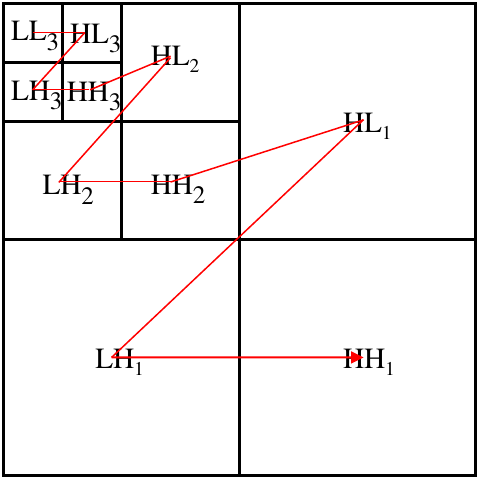} 
		~~~
		\includegraphics[width=0.4\columnwidth]{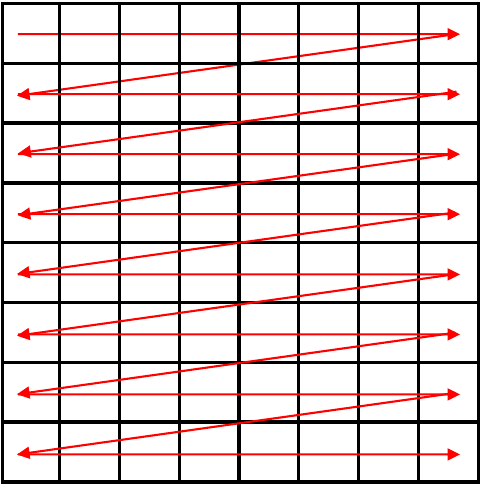} \\
		{\small (a) Order of inter-subbands}~~~~
		{\small (b) Order of intra-subband}
	\end{tabular}
	
	\caption{(a) The subbands are obtained by 3 level wavelet transform. The order of coding the subbands is denoted by the red line. (b) For each subband, the wavelet coefficients are coded one by one according to the red broken line.}
	\label{entropy-coding-order}
\end{figure}

After forward transform and quantization, the entropy encoding module is used to encode the quantized subband coefficients into bits. 
Fig.~\ref{entropy-coding-order}(a) shows the subbands obtained by three-level wavelet-like transform, and these subbands are coded one by one according to the red broken line. Within each subband, the subband coefficients are coded one by one according to the red broken line in Fig.~\ref{entropy-coding-order}(b).
When coding a subband coefficient, the currently coded subband and the previously coded subbands are first sent to the context model to derive entropy parameters. Then these entropy parameters are fed into the probability model to obtain a probability distribution. Finally, according to the probability distribution, the arithmetic coding engine codes the subband coefficients into binary bits. The following subsections introduce the context model and the probability model in detail.
\subsubsection{\textbf{Context Model}}
\label{ddd}
\begin{figure}
	\centering
	\includegraphics[width=1\columnwidth]{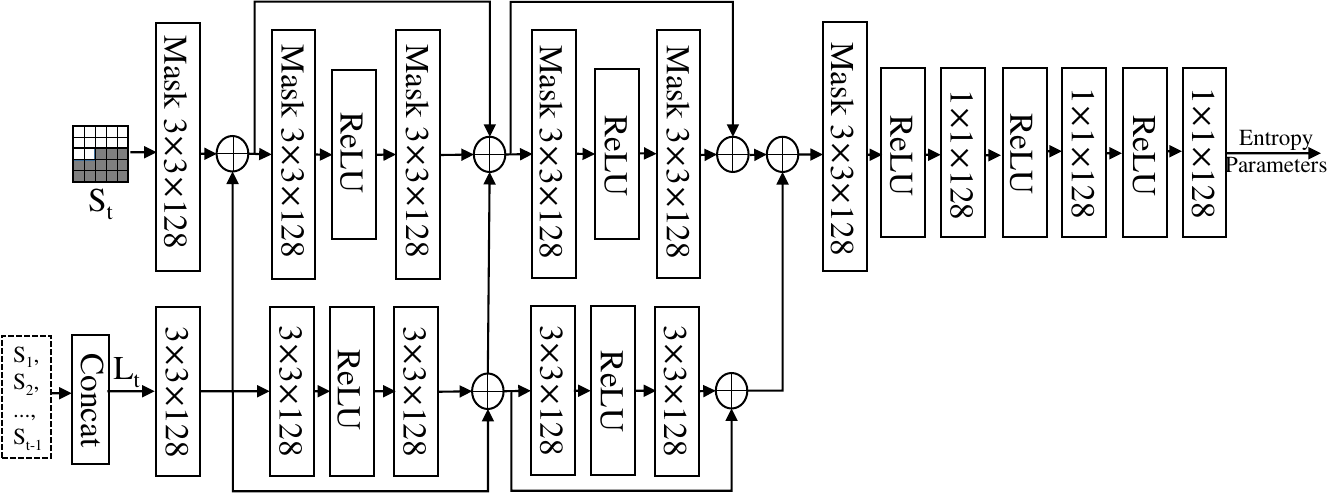} \\
	\caption{The structure of context model. The mask convolution \cite{van2016conditional} is used, as the $S_t$ has a partial area unavailable.}
	\label{context-fusion-model}
\end{figure}
As shown in Fig.~\ref{context-fusion-model}, the proposed context model is designed to extract two distinct types of contextual information, namely short-term context $S_t$ and long-term context $L_t$. 
The $S_t$ extracts the contextual information pertaining to the previously coded coefficients within the current subband.
The $L_t$ extracts the contextual information pertaining to the previously coded subbands and can be obtained through the concatenation of previously coded subbands. 
For example, when encoding a subband coefficient in $LH_3$ of Fig.~\ref{entropy-coding-order}(a), $LH_3$ is utilized as $S_t$, while $LL_3$ and $HL_3$ are concatenated to form $L_t$.  
After obtaining the short-term context $S_t$ and the long-term context $L_t$, they are fed into the network to derive the entropy parameters. 
In the context model, the mask convolution \cite{van2016conditional} is applied to $S_t$ for the causality.
A minor tweak worth noting is that in the case where the resolution of the current subband is different from that of the previous subbands, e.g. when coding $HL_2$ in Fig. \ref{entropy-coding-order}, we perform inverse wavelet transform on the previously coded \{$LL_3$, $HL_3$, $LH_3$, $HH_3$\} to obtain the long-term context $LL_2$ whose resolution is the same as $HL_2$.

\subsubsection{\textbf{Probability Model}}
The entropy parameters, which are the output of the context model, are fed into the probability model to determine the probability distribution. 
The probability model in this paper is a discretized Gaussian mixture likelihood model \cite{cheng2020learned}, formulated as
\begin{equation}
	P({\hat y}) =  \sum_{k=1}^{K}w_{k} \left(\int _{\hat y - \frac{1}{2}} ^{\hat y + \frac{1}{2}} \mathcal{N}(u_{k},\sigma^{2}_k)(y) ~dy\right)
	\label{partitions_new}
\end{equation}
where $\hat y$ denotes a quantized coefficient, $k$ denotes the index of mixtures, $K$ denotes the number of mixtures. Each mixture is characterized by a Gaussian distribution with 3 parameters, i.e., weight $w_{k}$, mean $u_{k}$, and variance $\sigma^{2}_k$, which are exactly the entropy parameters from the context model. 

\subsection{Post-Processing}
Since the inverse wavelet-like transform is designed as a perfect inversion of the forward transform, the distortion caused by quantization cannot be effectively handled by the decoder. 
Therefore, for lossy compression, we add a post-processing module to handle the distortion caused by quantization and improve the qualities of reconstructed images. 
The post-processing module consists of two submodules: dequantization and perceptual post-processing. For objective-oriented lossy coding, only dequantization is used, while both submodules are used for perceptual-oriented lossy coding. The rationale for using two submodules in perceptual-oriented lossy coding is to initially employ the dequantization module to minimize signal distortions, and subsequently apply the perceptual post-processing module, which is trained by GAN loss, to further enhance perceptual qualities of reconstructed images.

\begin{figure}
	\centering
	\includegraphics[width=0.7\columnwidth]{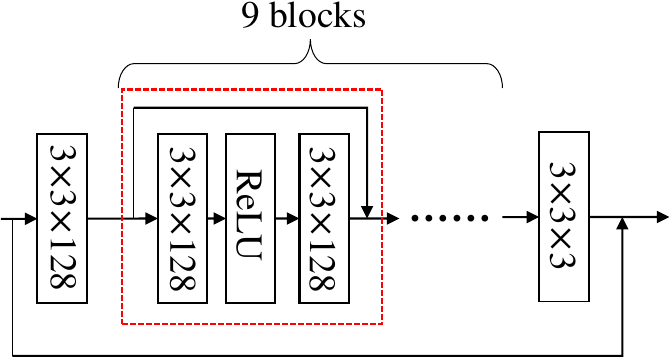} \\
	\caption{The network structure of perceptual post-processing.  }
	\label{gan-post}
\end{figure}

We design the dequantization module based on the RCAN architecture proposed by Zhang $et$ $al$.\cite{zhang2018image} but remove the upsampling layers of RCAN.
For the modified RCAN, we employ 10 residual groups, each has 10 residual blocks, and the number of feature channels of convolutional layers is set to 32.
The perceptual post-processing is a plain CNN and the network structure is shown in Fig. \ref{gan-post}. The network uses a global shortcut as well as multiple residual blocks, stabilizing and speeding up the training process.
\subsection{Training}
iWaveV3 supports three different coding configurations: 
objective-oriented lossy coding, perceptual-oriented lossy coding, and lossless coding. Each configuration requires a distinct training strategy. The following subsections provide detailed introductions to the training strategies of all coding configurations.
\subsubsection{\textbf{Objective-Oriented Lossy Coding}}
\label{Lossy coding for objective quality}

To achieve a high compression efficiency, we propose a three-step training approach for objective-oriented lossy coding. 
Considering that the context model and dequantization are more complex than the other modules, we propose to train them solely first. 
After that, we conduct an end-to-end training for the whole framework by using the soft-to-hard quantization technique \cite{agustsson2020universally}. 
Finally, to eliminate the quantization mismatch between training and test, the soft-then-hard quantization technique \cite{guo2021soft} is used to train the framework. 
Note that our training process supports the variable bitrate coding.

First, to train the entropy coding and the dequantization modules, we take the CDF 9/7 wavelet transform \cite{guangjun2001simple} and a fixed QStep as transform and quantization modules, respectively. 
Since the CDF 9/7 wavelet and the QStep are fixed, the context model module is trained to minimize the bitrate, while the dequantization module is trained to minimize MSE.

Second, with the pre-trained context model and dequantization modules as initialization, a CNN-based wavelet-like transform is introduced in place of the CDF 9/7 wavelet and a trainable QStep is introduced in place of the fixed QStep, enabling the end-to-end training of the whole framework.
The loss function for end-to-end training is as follows:
\begin{equation}
	\begin{aligned}
		loss = bpp + \lambda \times L_{obj}
	\end{aligned}
	\label{obj-loss} 
\end{equation}
where $\lambda$ is used to adjust the bitrate.
The distortion $L_{obj}$ is computed as:
\begin{equation}
	L_{obj} = \left \| I - \hat{I} \right\|_F
	\label{Lmse}
\end{equation}
where $I$ and $\hat{I}$ represent the original image and the reconstructed image, respectively.

Note that end-to-end training can be achieved only when the quantization is differentiable. 
Therefore, differentiable quantization techniques \cite{bengio2013estimating, theis2017lossy, balle2016end} are usually introduced to simulate the non-differentiable rounding quantization.
In this paper, we introduce the soft-to-hard quantization\cite{agustsson2020universally} to address this problem,
\begin{equation}
	\begin{aligned}
		&\lfloor y \rceil \approx s_\alpha(s_\alpha(y)+u) \\
		&s_\alpha(y) =  \lfloor y \rfloor + \frac{1}{2}\frac{tanh(\alpha r)}{tanh(\alpha / 2)} + \frac{1}{2} \\ 
		&r = y - \lfloor y \rfloor - \frac{1}{2}
	\end{aligned}
	\label{soft-to-hard} 
\end{equation}
where y and $\lfloor y \rceil$ denote the values before and after quantization, respectively, $u$ represents uniform noise ranging from -0.5 to 0.5, and $\alpha$ controls the fidelity of approximation to rounding. The smaller $\alpha$ means the closer approximation of quantization to uniform noise while the larger $\alpha$ makes the quantization closer to rounding. To prevent unstable training, an upper limit is set for $\alpha$. In our experiments, we set the initial value of $\alpha$ as 2, and it gradually anneals to 12.

Third, we propose to use the soft-then-hard quantization\cite{guo2021soft} strategy to further improve compression performance by eliminating the quantization mismatch between training and test.
In the soft-then-hard quantization method, the network parameters before quantization are fixed and the hard rounding is used in forward propagation. 
The loss function of training is the same as (\ref{obj-loss}).

Variable bitrate coding is an important feature of image compression.
However, in most end-to-end image coding frameworks, one model can only achieve a single bitrate. 
Our framework allows for multiple bitrates to be achieved using a single model through QStep adjustments.
The bitrate can be reduced by adding a positive offset to the trained QStep, and vice versa.
However, without considering the QStep adjustment during training, a significant performance drop will be suffered if the QStep is changed in the test phase.  
To address this problem, we take QStep adjustment into account by adding a random offset to the base QStep during the third training step (soft-then-hard). The random offset equips the trained network with the adaptation ability of QStep adjustment in the test phase.

\subsubsection{\textbf{Perceptual-Oriented Lossy Coding}}
The training of perceptual-oriented lossy coding is divided into two steps. We first train the whole network in Fig. \ref{overview} except for the perceptual post-processing module with the perceptual loss. Then the perceptual post-processing module is trained solely. The training details are described below.

First, we initialize the network parameters of perceptual-oriented lossy coding with that of the model trained in Section \ref{Lossy coding for objective quality}, considering that network structures are the same.
Then, the network is end-to-end trained through the same strategy used in the second (soft-to-hard) and third (soft-then-hard) steps in Section \ref{Lossy coding for objective quality}, but the loss function is changed. In addition to (\ref{obj-loss}), we add a feature loss, which has been proved to be friendly to the perceptual quality\cite{johnson2016perceptual, wang2018esrgan}, to further improve perceptual quality.
In this paper, we use the feature map of the pre-trained VGG-19\cite{simonyan2014very} network before the activation layer to calculate the feature loss,
\begin{equation}
	L_{vgg} = \left \| f_{vgg}(I) - f_{vgg}(\hat{I})) \right\|_F
	\label{Lvgg}
\end{equation}
where $f_{vgg}()$ represents the feature maps obtained from VGG-19. The overall loss function for end-to-end training is as follows:
\begin{equation}
	loss = bpp + \lambda \times (L_{obj} + \alpha \times L_{vgg})
	\label{loss-mse-vgg}
\end{equation}

\begin{figure}
	\centering
	\includegraphics[width=0.6\columnwidth]{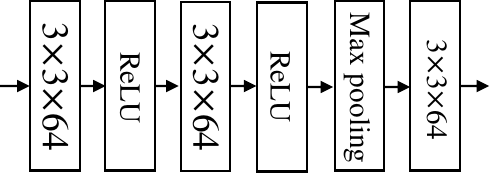} \\
	\caption{The network structure of discriminator network. The kernel size and stride of the max pooling are (2 $\times$ 2) and 2, respectively. }
	\label{D-network}
\end{figure}
Second, given that adversarial loss can constrain the distribution between reconstructed images and original images, we apply it to our perceptual post-processing module to improve the perceptual qualities of reconstructed images. We use the generative adversarial training method proposed in \cite{ma2022rectified}, which is easy to train and performs well. Specifically, the generator loss and the discriminator loss are as follows:
\begin{equation}
	\begin{aligned}
		&L_{G} = \frac{1}{2} \times E[ \left| D(I) - D(G(\hat I)) \right|]~+ \\ &~~~~~~~~\frac{1}{2} \times E[D(I) - D(G(\hat I))] \\ 
		&L_{D} = E[D(I) - D(G(\hat I))]
	\end{aligned}
	\label{gan}
\end{equation}
where the $G$ and $D$ are the perceptual post-processing network and the discriminator network, respectively. Their network structures are all plain CNNs and are shown in Fig. \ref{gan-post} and Fig. \ref{D-network}, respectively. 
In addition to GAN loss, we also add pixel domain loss and feature loss to stabilize training.
The overall loss function for the perceptual post-processing is as follows:
\begin{equation}
	loss = \alpha \times L_{obj} + \beta \times  L_{vgg} + \gamma \times L_G
	\label{loss-mse-vgg-gan}
\end{equation}
where $\alpha$, $\beta$, and $\gamma$ are hyperparameters used to control the weights of the three losses. Since the perceptual post-processing module does not change the bitrate, there is no bitrate in the loss function. Perceptual-oriented lossy coding \cite{mentzer2020high, agustsson2019generative} typically incorporates the adversarial loss during end-to-end training rather than using perceptual post-processing. However, in our experiments, we found that adding adversarial loss during end-to-end training can result in unstable training. Therefore, we have opted to use a separate perceptual post-processing.

\subsubsection{\textbf{Lossless Coding}}
For lossless coding, the quantization and post-processing modules in Fig. \ref{overview} are removed. Since no quantization module is required, the network can be trained end-to-end without concerning the gradient problem, and the loss function contains only the bitrate.
\subsection{Online Optimization}

An image coding framework trained on a dataset typically achieves the best compression performance on average within that dataset, but it is not optimized for a specific image. 
Optimizing for an individual image can improve compression performance, which is commonly referred to as online optimization \cite{wang2021substitutional, jiang2022online}. 
To further improve compression performance of the proposed framework, we introduce online optimization after offline training.
Similar to pre-processing, we update the image online and then code the updated image using the coding framework. We use the gradient descent method to update the image, with the specific formula as follows:
\begin{equation}
	\begin{aligned}
	 \widetilde{I} = I - grad*lr
	\end{aligned}
\end{equation}
where $I$ and $\widetilde{I}$ are the original image and the online updated image, respectively. $grad$ is the gradient, which is obtained through backpropagation, and $lr$ is the learning rate. Note that, the online optimization is only performed for objective-oriented lossy coding in this paper.

\section{Experiments}
\label{experment}
In this section, the experimental results are presented to validate the effectiveness of our method. Section \ref{setting} introduces the experimental settings. 
Sections \ref{objective-peoformence}, \ref{perceptual-peoformence} and \ref{lossless-peoformence} introduce the objective performance of lossy coding, the perceptual performance of lossy coding, and the compression ratio of lossless coding, respectively. Finally, the ablation study is presented in Section \ref{Ablation studies}.

\subsection{Expermental Settings}
\label{setting}
We use 1600 high-resolution images in the NIC dataset \footnote{https://structpku.github.io/LIU4K\_Dataset/LIU4K\_v2.html} and 800 high-resolution images in the DIV2K training set\cite{agustsson2017ntire} for training. The images are randomly cropped into blocks of size 128$\times$128. The blocks are used as training data without any data augmentation. The batch size is set to 12. For the framework with additive transform, the Adam optimizer\cite{kingma2014adam} with  $ lr = 1 \times 10^{-4}, \beta_1 = 0.9$ and $ \beta_2 = 0.999$ is used. For the framework with affine transform, the Adam optimizer\cite{kingma2014adam} with  $ lr = 1 \times 10^{-5}, \beta_1 = 0.9$ and $ \beta_2 = 0.999$ is used. For the perceptual post-processing module, RMSprop optimizer\cite{tieleman2012rmsprop} with  $ lr = 1 \times 10^{-5}, \alpha = 0.9$ and $\epsilon = 10^{-8}$ is used to train the perceptual post-processing network and RMSprop optimizer\cite{tieleman2012rmsprop} with $ lr = 5 \times 10^{-6}, \alpha = 0.9$ and $\epsilon = 10^{-8}$ is used to train the discriminator network. The perceptual post-processing network and discriminator network are optimized in an alternating manner, the discriminator network is optimized 5 times, and then the perceptual post-processing network is optimized once. 

In the testing stage, for the objective performance of lossy coding, we use peak signal-to-noise ratio (PSNR) and bitrate as evaluation metrics. For the perceptual performance of lossy coding, we use learned perceptual image patch similarity (LPIPS)\cite{zhang2018unreasonable} and bitrate as metrics. We also give reconstructed images for subjective comparison.
For lossless coding, we use compression ratio as a metric.

\subsection{The Performance of Objective Quality Lossy Coding}
\label{objective-peoformence}
\begin{figure}
	\centering
	\includegraphics[width=0.8\columnwidth]{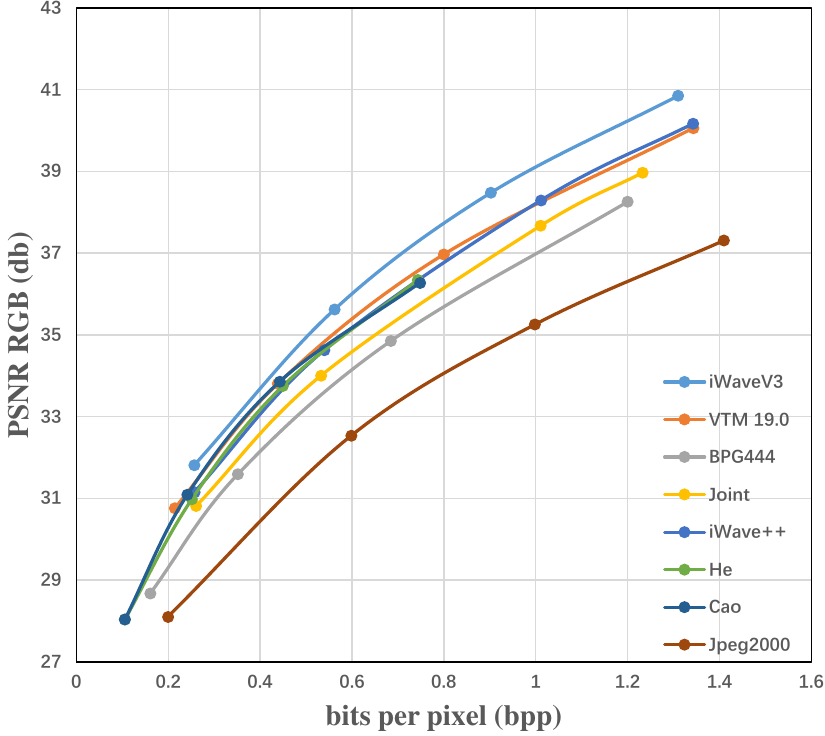} \\
	\caption{Comparison of different lossy compression methods on the
		Kodak dataset. The average rate (bits-per-pixel) and the
		average PSNR (calculated in RGB) of the 24 images  are shown. “Joint” refers
		to\cite{minnen2018joint}, “He” refers to\cite{he2021checkerboard}, and “Cao” refers to\cite{cao2022end}.}
	\label{kodak_psnr}
\end{figure}

In this subsection, we compare our objective-oriented lossy compression method iWaveV3 with several
typical lossy compression methods, including non-deep learning methods: JPEG-2000 (using the OpenJPEG software), BPG, VVC test model VTM19.0\footnote{https://vcgit.hhi.fraunhofer.de/jvet/VVCSoftware$\_$VTM} (4:4:4 format), and deep learning methods: Joint\cite{minnen2018joint}, iWave++\cite{ma2020end}, He\cite{he2021checkerboard} and Cao\cite{cao2022end}. All models are tested on two independent image datasets, i.e., Kodak (24 low-resolution images) \cite{kodak1993kodak} and Tecnick (100 high-resolution images) \cite{asuni2014testimages}, which are widely used as the test set for image compression.

\begin{figure}
	\centering
	\includegraphics[width=0.8\columnwidth]{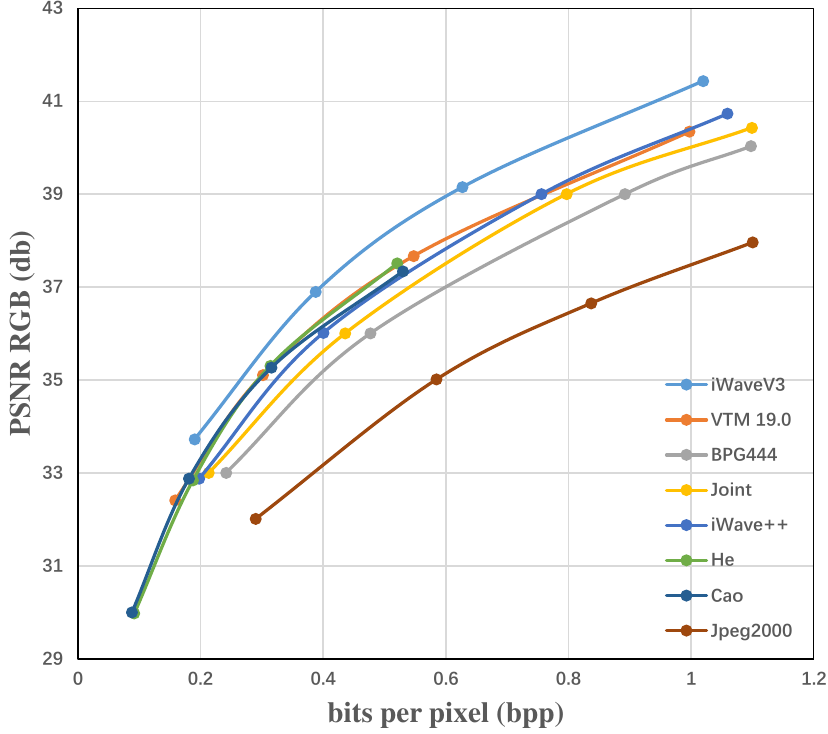} \\
	\caption{Comparison of different lossy compression methods on the
		Tecnick dataset. The average rate (bits-per-pixel) and the
		average PSNR (calculated in RGB) of the 100 images are shown. “Joint” refers
		to\cite{minnen2018joint}, “He” refers to\cite{he2021checkerboard}, and “Cao” refers to\cite{cao2022end}.}
	\label{tecnick_psnr}
\end{figure}

The results measured by PSNR are shown in Fig.~\ref{kodak_psnr}
and Fig.~\ref{tecnick_psnr} for Kodak and Tecnick test sets, respectively. iWaveV3 performs better than all other methods, including non-deep and deep methods, on both datasets. 
We also present quantitative performance in Tabel \ref{BD-rate}. Compared with iWave++, iWaveV3 improves compression performance by 13.5\% and 19.76\% on the Kodak and Tecnick datasets, respectively.  
Compared with VTM19.0, iWaveV3 can improve compression performance by 11.07\% and 15.64\% on the Kodak and Tecnick datasets, respectively, which demonstrates iWaveV3's excellent compression performance.
iWaveV3 has also been submitted to IEEE 1857.11 CfP, and the test set is 20 images with a resolution of 4k. Tabel \ref{cfp-BD-rate} shows the BD-rate reduction result compared with the BPG. Among all the methods, our method iWaveV3(Online) achieves the first place on the PSNR metric.
\begin{table}[]
	\caption{the BD-rate reduction of iWaveV3 when anchored on iWave++\cite{ma2020end} and VTM 19.0. }
	\begin{center}
		\begin{tabular}{c|c|c}
			\hline
			\multicolumn{1}{l|}{}              & dataset & BD-rate (\%)                \\ \hline
			\multirow{2}{*}{Anchored on iWave++} & Kodak   & -13.5\%                       \\ \cline{2-3} 
			& Tecnick & -19.76\%                      \\ \hline
			\multirow{2}{*}{Anchored on VTM19.0} & Kodak   & -11.07\%                      \\ \cline{2-3} 
			& Tecnick & \multicolumn{1}{c}{-15.64\%} \\ \hline
		\end{tabular}
		\label{BD-rate}
	\end{center}
\end{table}

\begin{table}[]
	\caption{the BD-rate reduction of various methods of 1857.11 CfP  anchored on BPG. }
	\begin{center}
	\begin{tabular}{c|c}
		\hline
		Method          & BD-rate(\%) \\ \hline
		VTM Intra       & -20.01\%    \\ \hline
		NIC (MSE)       & -31.08\%    \\ \hline
		NIC (MS-SSIM)   & -18.77\%    \\ \hline
		iWaveV3 (MSE)    & -33.54\%    \\ \hline
		iWaveV3 (Percep) & 14.23\%     \\ \hline
		{iWaveV3 (Online)} & {-36.80}\%    \\ \hline
	\end{tabular}
\end{center}
\label{cfp-BD-rate}
\end{table}

\subsection{The Performance of Perceptual Quality Lossy Coding}
\label{perceptual-peoformence}
In this subsection, we compare our perceptual-oriented lossy coding method iWaveV3-Perp with other methods, including VTM19.0, HiFiC\cite{mentzer2020high}, and our scheme iWaveV3-Obj optimized for objective quality. We test the performance of all methods in terms of LPIPS\cite{zhang2018unreasonable} which is widely used for perceptual quality evaluation. The Kodak and clic2020 are used as test sets, so as to be compared with HiFiC\cite{mentzer2020high}. Kodak has 24 low-resolution images, and clic2020 has 428 high-resolution images.

\begin{figure}
	\centering
	\includegraphics[width=0.8\columnwidth]{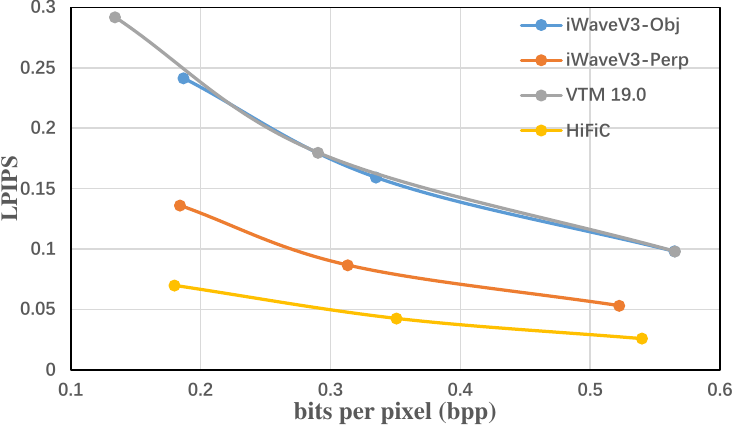} \\
	\caption{Subjective quality comparison of different compression methods on the
		Kodak dataset. The average rate (bits-per-pixel) and the
		average LPIPS of the 24 images are shown.}
	\label{kodak-lpips}
\end{figure}
\begin{figure}
	\centering
	\includegraphics[width=0.8\columnwidth]{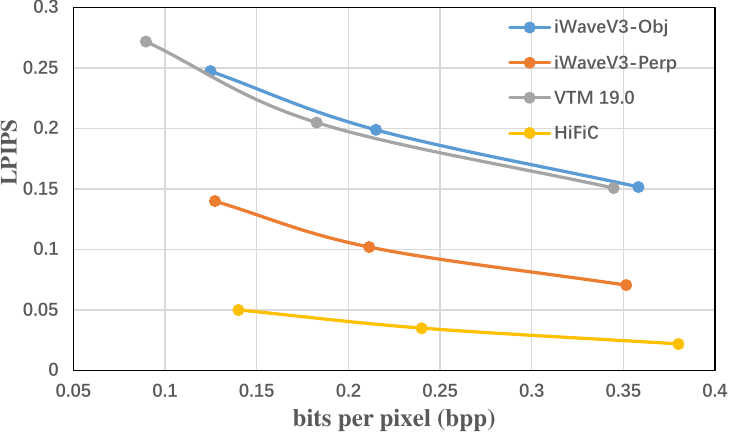} \\
	\caption{Subjective quality comparison of different compression methods on the
		clic2020 dataset. The average rate (bits-per-pixel) and the
		average LPIPS of the 428 images are shown.}
	\label{clic2020-lpips}
\end{figure}

The results measured by LPIPS are shown in Fig.~\ref{kodak-lpips}
and Fig.~\ref{clic2020-lpips} for Kodak and clic2020 test sets, respectively. It can be seen that the LPIPS of iWaveV3-Perp is significantly lower than that of iWaveV3-Obj, which shows that our improvements for perceptual quality do have a significant effect. The LPIPS of iWaveV3-Perp is also significantly better than the best traditional method VTM19.0. However, iWaveV3-Perp is slightly inferior compared to HiFiC. This may be due to the fact that the nonlinearity and flexibility of wavelet-like transform are slightly worse than that of the auto-encoder.

\begin{figure*}
	\centering
	\includegraphics[width=2\columnwidth]{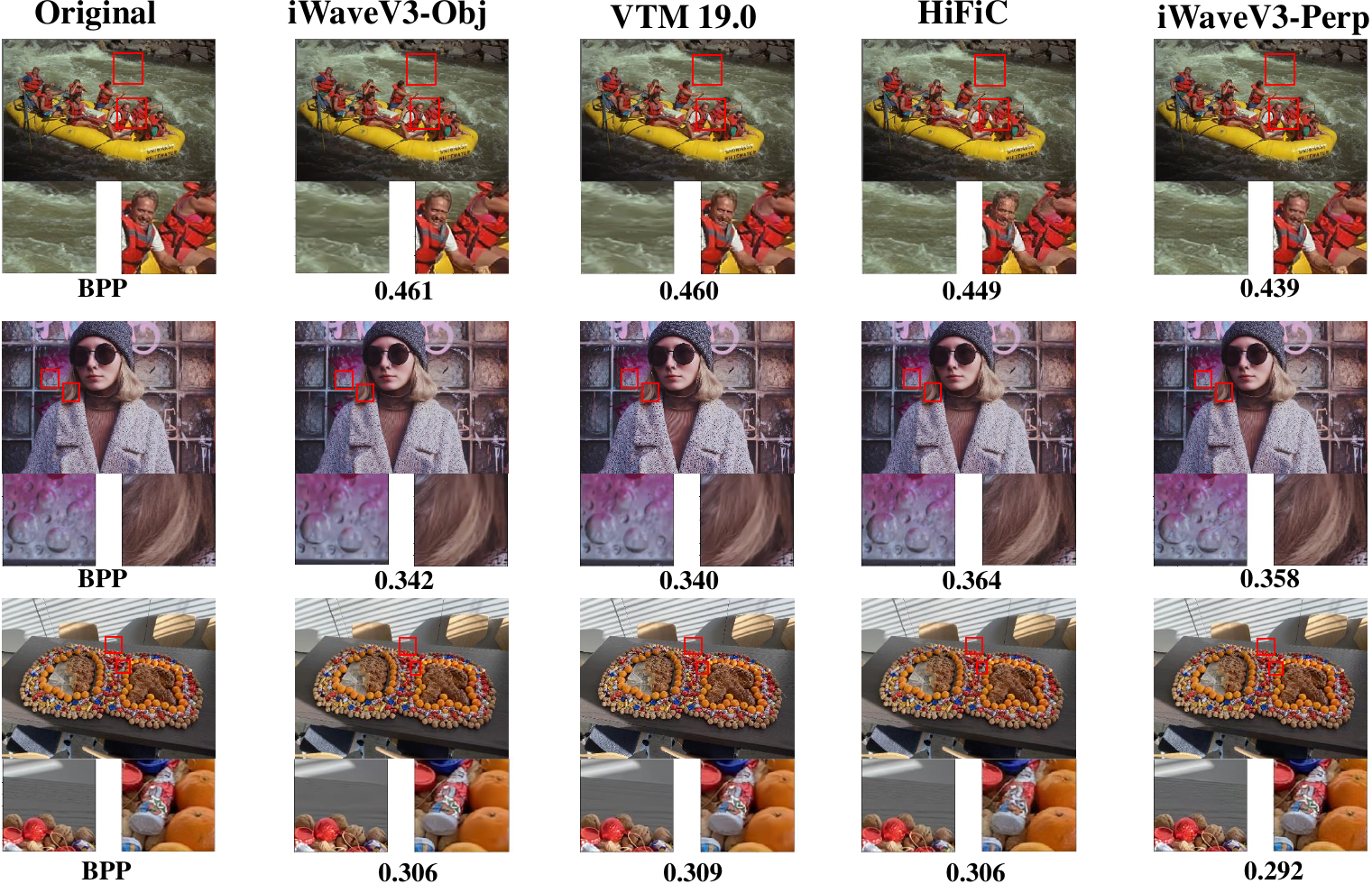} \\
	\caption{Visualization of reconstructed images by different compression methods from Kodak and clic2020 datasets.}
	\label{perceptual-image}
\end{figure*}
We show reconstructed images from Kodak and clic2020 for different methods in Fig. \ref{perceptual-image}. The graphs in the first row are from Kodak, and the graphs in the second and third rows are from clic2020. The reconstructed images of iWaveV3-Perp exhibit significant improvement over those of iWaveV3-Obj and the traditional method VTM19.0. They feature sharper edges and more intricate textures, making them visually superior. Compared with HiFiC, iWaveV3-Perp has better reconstruction results on complex textures, such as the human face in the first row and the candy in the third row. But the effect is not as good as HiFiC on simple textures, such as the hair in the second row and the lines in the third row. 

\subsection{The Performance of Lossless Coding}
\label{lossless-peoformence}
In this subsection, we test the compression performance of lossless coding. We compare iWaveV3 with JPEG-2000, FLIF\cite{sneyers2016flif}, WebP\footnote{http://code.google.com/speed/webp/} and iWave++. We tested them on the Kodak dataset, and the test results are shown in Tabel \ref{lossless}. It can be seen that compared with iWave++, iWaveV3 can achieve higher compression efficiency. We can also see that iWaveV3 is significantly better than JPEG-2000 and WebP, almost the same as FLIP. In addition, on the 1857.11 CfP test sets, the average bpp of BPG is 14.32, that of FLIF is 8.79, and that of our method is 8.54.
\begin{table}[]
	\caption{Comparison of different lossless compression methods on the Kodak
		dataset (bits per pixel)}
	\begin{center}
	\begin{tabular}{c|c|c|c|c|c}
		\hline
		Method & JPEG-2000 & WebP & FLIF & iWave++ & iWaveV3 \\ \hline
		Kodak  & 9.58      & 9.56 & 8.68 & 8.97    & 8.75      \\ \hline
	\end{tabular}
\end{center}
\label{lossless}
\end{table}

\subsection{Ablation Studies}
\label{Ablation studies}
iWaveV3 is based on iWave++, but has substantial new technologies compared with iWave++. In this subsection, we conduct ablation studies on these new technologies. First, we introduce technologies to improve objective compression performance: the affine wavelet-like transform, the quantization strategies, the entropy coding, the post-processing, the online optimization, and so on. Then we introduce the technologies made to adapt to the variable bitrate. Finally, we introduce the technologies for perceptual quality.

\begin{figure}
\centering
\includegraphics[width=0.8\columnwidth]{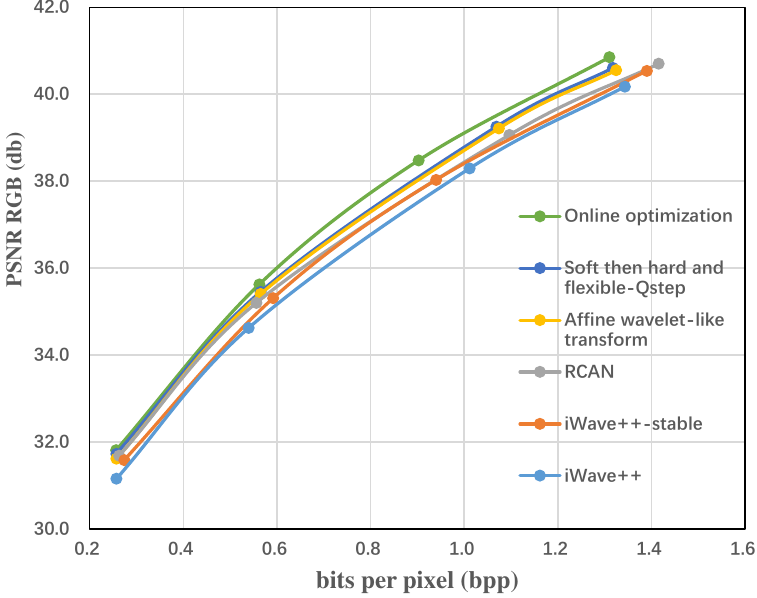} \\
\caption{The ablation studies of our framework tested on the Kodak dataset. iWave++-stable is the stable version of iWave++. On the basis of iWave++-stable, we gradually added four technologies, which are gray, yellow, blue, and green lines, respectively.}
\label{ablation_study}
\end{figure}

In order to improve the objective compression performance, we have introduced several new technologies. To confirm their effectiveness, we conduct ablation experiments in the following order.
First, in order to make iWave++ easier to train, we replace the STE quantization used in iWave++ with soft-to-hard quantization, replace the LSTM module with concatenate operation and replace the very heavy probability model with the discrete Gaussian mixture probability model. We record this version as iWave++-stable. Then on the basis of iWave++-stable, we replace the previous very simple dequantization module with the more advanced RCAN. Third, we add an affine wavelet-like transform. On this basis, the soft-then-hard technology and flexible-QStep for variable bitrate are used. Finally, we add online optimization technology. We perform ablation experiments on the Kodak dataset, and the experimental results are shown in Fig. \ref{ablation_study}.
We also give accurate performance gains compared with iWave++ by averaging BD-rate reduction, which is shown in Table~\ref{BD-rate-ablation_study}. 
\begin{table}[]
	\caption{the BD-rate reduction that our method achieve when anchored on iWave++\cite{ma2020end} on Kodad dataset.}
	\renewcommand{\baselinestretch}{1}\footnotesize
	\resizebox{\columnwidth}{!}{
	\begin{tabular}{c|c|c|c|c|c}
		\hline
		Stable & RCAN & \begin{tabular}[c]{@{}c@{}}Affine\\      wavelet-like\end{tabular} & \begin{tabular}[c]{@{}c@{}}Soft then hard \\ and\\      flexible-QStep\end{tabular} & \begin{tabular}[c]{@{}c@{}}Online \\      optimization\end{tabular} & \begin{tabular}[c]{@{}c@{}}BD-rate\\      reduction\end{tabular}
		      \\ \hline
		\CheckmarkBold           & \XSolidBrush & \XSolidBrush                                                                & \XSolidBrush                                                                                & \XSolidBrush                                                            & -3.25\%  \\ \hline
		\CheckmarkBold           & \CheckmarkBold & \XSolidBrush                                                                & \XSolidBrush                                                                                & \XSolidBrush                                                            & -6.61\%  \\ \hline
		\CheckmarkBold           & \CheckmarkBold & \CheckmarkBold                                                                & \XSolidBrush                                                                                & \XSolidBrush                                                            & -9.16\%  \\ \hline
		\CheckmarkBold           & \CheckmarkBold & \CheckmarkBold                                                                & \CheckmarkBold                                                                                & \XSolidBrush                                                            & -10.28\% \\ \hline
		\CheckmarkBold           & \CheckmarkBold & \CheckmarkBold                                                                & \CheckmarkBold                                                                                & \CheckmarkBold                                                            & -13.50\% \\ \hline
	\end{tabular}
}
\label{BD-rate-ablation_study}
\end{table}

\begin{figure}
	\centering
	\includegraphics[width=0.8\columnwidth]{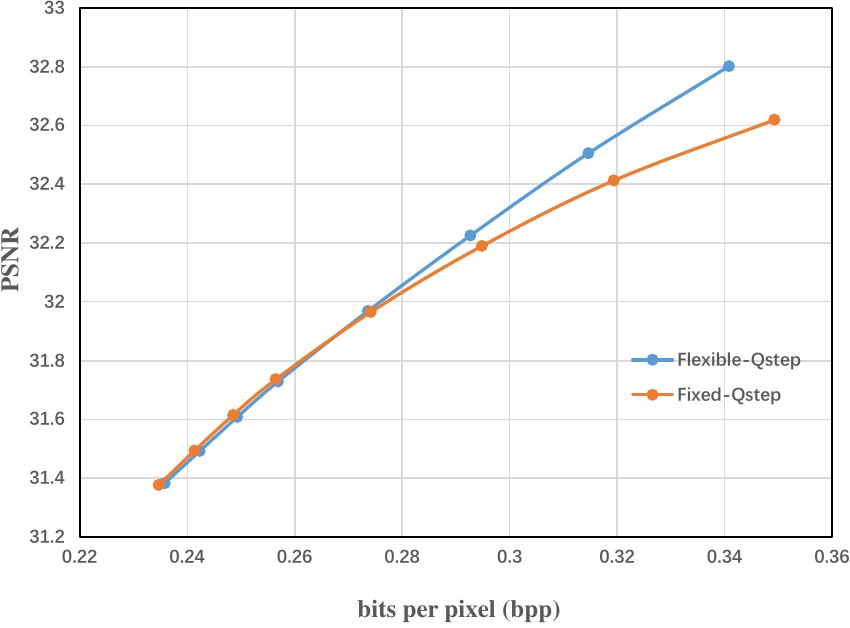} \\
	\caption{The R-D curves of a fixed-QStep model and a flexible-QStep model obtained by adjusting QStep in test phase. The average rate (bits-per-pixel) and the average PSNR (calculated in RGB) of Kodak dataset are shown.}
	\label{vair-rate}
\end{figure}
Our method can achieve variable bitrate by adjusting QStep. When QStep increases, the bitrate decreases, and vice versa. 
Fig. \ref{vair-rate} shows the RD curves of a fixed-QStep model and a flexible-QStep model. 
In the fixed-QStep model, the QStep is fixed during training while it is randomly adjusted by an offset in the flexible-QStep model. 
As shown in Fig. \ref{vair-rate}, the flexible-QStep model achieves significantly better performance than the fixed-QStep model when the bitrate increases.

\begin{figure}
	\centering
	\includegraphics[width=0.9\columnwidth]{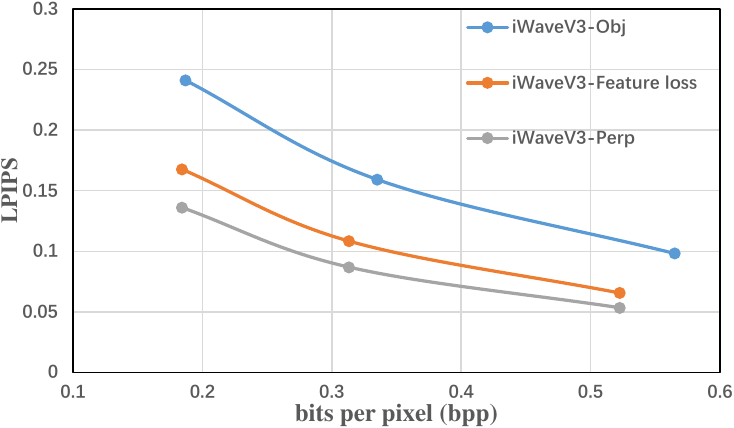} \\
	\caption{The ablation studies of our perceptual quality scheme tested on the Kodak dataset. The average rate (bits-per-pixel) and the
		average LPIPS of the 24 images are shown.}
	\label{ablation_study-percep}
\end{figure}

We propose to use feature loss and a perceptual post-processing network to further improve perceptual quality.
The Kodak dataset is used to evaluate the performance of the two technologies, and the RD curves are shown in Fig. \ref{ablation_study-percep}. 
As shown in Fig. \ref{ablation_study-percep}, compared with objective-oriented lossy coding iWaveV3-Obj, the iWaveV3 with feature loss has a significant improvement on LPIPS.
Moreover, the addition of the perceptual post-processing module, which is named as iWaveV3-Perp, achieves a better performance, as shown in Fig. \ref{ablation_study-percep}.

\section{Conclusion}
\label{conclusion}

In this paper, we present an end-to-end wavelet-like transform-based image coding framework iWaveV3, which simultaneously offers lossless coding, 
objective-oriented lossy coding, and perceptual-oriented lossy coding. The framework comprises various components, such as transform, quantization, entropy coding, and post-processing modules.
The transform module offers a choice of additive wavelet-like transform, affine wavelet-like transform, or CDF 5/3 wavelet transform.
The quantization module initially divides the subbands by the QStep and then quantizes them to integers by rounding. 
The entropy coding module is composed of a CNN-based context model, a probability model, and an arithmetic coding engine. The post-processing module is utilized to reduce the quantization distortion and enhance reconstruction quality.
Additionally, we also use advanced offline training and online optimization strategies to improve compression performance and use the perceptual metric to improve the perceptual qualities of reconstructed images.
Experimental results demonstrate that iWaveV3's compression performance is highly competitive with state-of-the-art image coding frameworks. As a result, it is chosen as a candidate scheme for the development of the IEEE Standard for Neural Network-Based Image Coding.
\bibliographystyle{IEEEtran}
\bibliography{iWave-CSVT}

\vfill

\end{document}